\documentclass[pdflatex,sn-nature]{sn-jnl}% Math and Physical Sciences Numbered Reference Style
\usepackage{graphicx}%
\usepackage{multirow}%
\usepackage{amsmath,amssymb,amsfonts}%
\usepackage{amsthm}%
\usepackage{mathrsfs}%
\usepackage[title]{appendix}%
\usepackage{xcolor}%
\usepackage{textcomp}%
\usepackage{manyfoot}%
\usepackage{booktabs}%
\usepackage{algorithm}%
\usepackage{algorithmicx}%
\usepackage{algpseudocode}%
\usepackage{listings}%
\theoremstyle{thmstyleone}%
\theoremstyle{thmstyletwo}%

\theoremstyle{thmstylethree}%

\begin{document}

\title[Article Title]{Frequency-aware forecasting for short-term typhoon gust prediction}
%%=============================================================%%
%% GivenName	-> \fnm{Joergen W.}
%% Particle	-> \spfx{van der} -> surname prefix
%% FamilyName	-> \sur{Ploeg}
%% Suffix	-> \sfx{IV}
%% \author*[1,2]{\fnm{Joergen W.} \spfx{van der} \sur{Ploeg} 
%%  \sfx{IV}}\email{iauthor@gmail.com}
%%=============================================================%%
\author[1]{\fnm{Xuefei} \sur{Wang}}\email{xuefeiw99@gmail.com}

\author[2]{\fnm{Tingyi} \sur{Liu}}\email{tingyiL@whu.edu.cn}

\author[3]{\fnm{Heng} \sur{Zhang}}\email{zhangheng\_sjtu@sjtu.edu.cn}

\author*[4]{\fnm{Lei} \sur{Xu}}\email{lei5@kth.se}

\author*[1]{\fnm{Shengjun} \sur{Zhang}}\email{sj.zhang@hubu.edu.cn}

\affil[1]{\orgdiv{School of Artificial Intelligence}, 
\orgname{Hubei University}, 
\orgaddress{\city{Wuhan}, \country{China}}}

\affil[2]{\orgdiv{School of Economics and Management}, 
\orgname{Wuhan University}, 
\orgaddress{\city{Wuhan}, \country{China}}}

\affil[3]{\orgdiv{School of Electrical Engineering}, 
\orgname{Shanghai Jiao Tong University}, 
\orgaddress{\city{Shanghai}, \country{China}}}

\affil[4]{\orgdiv{Division of Decision and Control Systems, School of Electrical Engineering and Computer Science}, 
\orgname{KTH Royal Institute of Technology}, 
\orgaddress{\postcode{100 44}, \city{Stockholm}, \country{Sweden}}}
%%==================================%%
%% Sample for unstructured abstract %%
%%==================================%%

\abstract{Accurate gust forecasting under typhoon conditions remains challenging due to the highly non-stationary and multi-scale characteristics of extreme wind fluctuations. Existing deep learning models often struggle to simultaneously capture long-term trends and rapid local variations, resulting in degraded performance during extreme events. We propose WDANet, a frequency-aware forecasting framework that integrates stationary wavelet decomposition, a Feature-wise Linear Modulation (FiLM) strategy, and a dual-branch encoder–decoder architecture, enabling separate modeling of trend and fluctuation components. Taking the offshore regions of the Western Pacific in China as an example, we conduct fine-grid wind gust prediction research. The results demonstrate that WDANet shows advantages for short lead times under the experimental setting across a 24-h forecasting horizon and achieves higher prediction accuracy than ECMWF-HRES within the first 6 h. During extreme wind events, WDANet more accurately captures gust peaks and attains the best RMSE and MAE performance. These results highlight its potential for offshore wind power operation, disaster warning, and risk mitigation. }

\maketitle

\section{Introduction}\label{sec1}

Offshore wind energy has emerged as a key pathway toward global carbon neutrality \cite{gielen2021world}, with cumulative installed capacity projected to increase from 63 GW in 2022 to nearly 494 GW by 2030 \cite{alliance2023tripling}. Despite its substantial offshore wind potential \cite{xu2026substantially}, a large fraction of China's offshore wind farms are located in the Northwest Pacific\cite{wang2024spatial,ou2018offshore}, one of the most active tropical cyclone basins worldwide. In recent years, the frequency and intensity of typhoons in this region have shown an increasing trend \cite{zhao2018changes,li2023recent,zhao2025increasing}, posing increasing challenges to the climate resilience and safe operation of offshore wind infrastructure.

During typhoon events, offshore wind turbines are exposed not only to elevated mean wind speeds but also to intense gusts \cite{worsnop2017gusts}. Unlike mean wind speed, which represents temporally averaged atmospheric conditions, gusts correspond to short-duration wind peaks that impose significantly larger aerodynamic loads on turbine blades, towers, and support structures\cite{onol2017effects,wu2019effects}. Excessive gusts may trigger emergency shutdown procedures, accelerate structural fatigue, and in severe cases cause catastrophic failures\cite{hou2022review,ning2025assessment}. Consequently, operational safety and risk management in offshore wind farms depend more directly on gust characteristics than on mean wind speed alone. 

However, accurate gust forecasting remains substantially more challenging than mean wind speed prediction. Mean wind speed evolves relatively smoothly and is dominated by large-scale atmospheric processes, whereas gusts arise from localized turbulence, convection, and rapid energy transfer across multiple temporal scales \cite{beljaars1987influence}. Under typhoon conditions, gust behavior exhibits strong intermittency, abrupt transitions, and pronounced non-stationarity \cite{worsnop2017gusts}. These characteristics make gust dynamics considerably more difficult to represent and predict using conventional forecasting models.

Despite their operational importance, most existing forecasting studies focus primarily on mean wind speed rather than gusts \cite{wang2020probabilistic,zhang2025optimization}. Conventional forecasting targets are typically temporally averaged quantities that suppress short-term extreme fluctuations. As a result, models optimized for mean wind speed often fail to accurately represent transient gust peaks, creating a mismatch between forecasting objectives and the variables that govern turbine safety under typhoon conditions.

Wind speed forecasting methods can be broadly categorized into physics-based and data-driven approaches \cite{wang2021review}. Numerical weather prediction (NWP) models provide physically interpretable forecasts but suffer from high computational cost and limited spatial-temporal resolution \cite{bauer2015quiet}. Data-driven approaches, including statistical models and deep learning architectures such as LSTM \cite{graves2012long}, GRU \cite{cho2014learning}, and Transformer-based models like Informer \cite{zhou2021informer} and SIGMAformer \cite{kim2026sigmaformer}, have significantly improved predictive performance by capturing temporal dependencies. Nevertheless, despite their improved capability in modeling long-term temporal dependencies, existing deep learning approaches still struggle to accurately represent sudden and high-frequency wind variations under highly non-stationary typhoon conditions. This limitation is primarily attributed to the complex multi-scale nature of typhoon-driven wind fields, where low-frequency atmospheric evolution and high-frequency turbulent gusts are strongly coupled and exhibit highly intermittent behavior \cite{gallego2015review}. As a result, existing models tend to smooth out extreme short-term fluctuations, leading to suboptimal representation of gust dynamics that are critical for offshore wind turbine safety.

To address multi-scale dynamics, a common strategy is to apply signal decomposition techniques such as discrete wavelet transform (DWT) or empirical wavelet transform (EWT) prior to forecasting \cite{liu2018wind,jiajun2020ultra}. These approaches assume that decomposed sub-series within specific frequency bands exhibit simpler temporal patterns \cite{stephane1999wavelet}. However, many existing decomposition-based methods rely on downsampling operations or non-shift-invariant filtering, which may introduce two potential limitations: (i) reduced temporal resolution, and (ii) sensitivity to time shifts that can affect the alignment of transient gust events. These methods often treat decomposed frequency components as relatively independent forecasting targets, which may overlook the cross-scale interactions between low-frequency atmospheric evolution and high-frequency gust fluctuations.

To overcome these limitations, this work introduces a Wavelet Dual-branch Attention Network (WDANet) for gust forecasting. Unlike traditional wavelet-based methods, SWT is fully non-decimated and shift-invariant, preserving temporal resolution across all decomposition levels. This property ensures that transient gust structures remain temporally aligned throughout the decomposition process, which is essential for accurately capturing abrupt wind changes.

Within WDANet, wind speed series are decomposed into low-frequency components capturing smooth atmospheric evolution and high-frequency components representing rapid gust-induced fluctuations. Furthermore, we design a frequency-aware learning strategy. The low-frequency component is used to model long-term atmospheric trends, while the high-frequency component is reformulated as a residual dynamic process emphasizing transient gust deviations. Considering that high-frequency gust bursts are not independent stochastic noise but are physically modulated by the evolving low-frequency typhoon background field, we introduce Feature-wise Linear Modulation (FiLM) \cite{perez2018film}. FiLM explicitly models the dependency between large-scale typhoon evolution and small-scale gust fluctuations by dynamically adjusting feature representations conditioned on the low-frequency atmospheric state. This allows a physically consistent integration of multi-scale wind dynamics.

The main contributions of this work are summarized as follows:
\begin{itemize}
\item We identify that temporal misalignment, loss of shift-invariance, and independence assumptions in decomposition-based methods are key bottlenecks for gust-level wind forecasting under typhoon conditions.
\item We propose a shift-invariant SWT-based multi-resolution forecasting framework that preserves full temporal resolution and maintains temporal coherence of transient gust structures.
\item We introduce a frequency-aware learning strategy that explicitly decouples smooth atmospheric evolution and high-frequency gust dynamics in a physically consistent manner.
\item The proposed method achieves superior gust forecasting accuracy against other machine learning models and outperforms ECMWF-HRES in the short term.
\end{itemize}

\section{Results}\label{sec2}

\subsection{Evaluation metrics}\label{sec:Evaluation metrics}

To comprehensively evaluate the forecasting performance, three complementary metrics are adopted in this work, including the root mean squared error (RMSE), mean absolute error (MAE), and relative accuracy (RAcc). RMSE and MAE are commonly used to quantify the prediction errors in wind gust forecasting, and their definitions are given as follows:

\begin{equation}
\label{RMSE}
RMSE = \sqrt{\frac{1}{N}\sum_{j=1}^{N}(G_j - \hat{G}_j)^2}
\end{equation}

\begin{equation}
\label{MAE}
MAE = \frac{1}{N}\sum_{j=1}^{N}\left|G_j - \hat{G}_j\right|
\end{equation}

where $N$ denotes the number of samples in the test set, $G_j$ represents the observed wind gust value of the $j$-th sample, and $\hat{G}_j$ denotes the corresponding predicted wind gust value. RMSE assigns larger penalties to samples with greater prediction deviations, making it more sensitive to extreme forecasting errors and abrupt gust events. In contrast, MAE measures the average absolute deviation over all samples, providing a more robust and interpretable evaluation of the overall forecasting accuracy.

Although RMSE and MAE are effective for quantifying absolute prediction errors, they cannot reflect the relative forecasting accuracy under different wind speed conditions. Considering the significant variability of wind gust intensity during typhoon evolution, evaluating the normalized prediction error is essential for comparing forecasting performance across different wind speed levels. Therefore, the relative accuracy (RAcc), derived from the mean absolute percentage error (MAPE), is further introduced:

\begin{equation}
\label{RAcc}
RAcc = 1 - MAPE = 1 - \frac{1}{N}\sum_{j=1}^{N}\left|\frac{G_j-\hat{G}_j}{G_j}\right|
\end{equation}

RAcc evaluates the forecasting performance by measuring the prediction deviation relative to the ERA5 reference gust, thereby providing a normalized assessment across different wind speed regimes. A higher RAcc value indicates that the predicted wind gust is closer to the observation, demonstrating stronger relative forecasting capability and better model generalization.

Furthermore, to provide a more detailed evaluation of the model capability in capturing extreme wind gust events, two additional metrics, namely peak error (PE) and peak time error (PTE), are introduced. PE quantifies the amplitude discrepancy between the predicted and observed peak gust values, while PTE measures the temporal shift between the predicted and observed peak occurrence times. They are defined as follows:

\begin{equation}
\label{PE}
PE = \left| G_{peak} - \hat{G}_{peak} \right|
\end{equation}

\begin{equation}
\label{PTE}
PTE = \left| t_{peak} - \hat{t}_{peak} \right|
\end{equation}

where $G_{peak}$ and $\hat{G}_{peak}$ represent the maximum wind gust values in the observed and predicted sequences, respectively. $t_{peak}$ and $\hat{t}_{peak}$ denote the corresponding time indices of the observed and predicted peak gusts. PE directly characterizes the accuracy of extreme gust intensity prediction, whereas PTE evaluates the capability of the model to accurately capture the arrival time and occurrence moment of sudden gust events. These two metrics complement the aforementioned global error measurements and provide essential evidence for assessing the practical applicability and reliability of the proposed model under extreme weather conditions.

\subsection{Comparison methods analysis}\label{sec:Comparison methods analysis}

To validate the effectiveness of the proposed WDANet framework for wind gust forecasting, several representative models with different architectures are selected for comparison on the test dataset. The details of these baseline methods are summarized as follows:

\begin{quote}
CNN-LSTM: A hybrid architecture that employs convolutional layers to extract local temporal features and integrates LSTM networks to model sequential dependencies, thereby capturing the temporal dynamics of wind gust evolution \cite{zhang2021wind}.
\end{quote}

\begin{quote}
Autoformer: A Transformer-based architecture that utilizes series decomposition and an autocorrelation mechanism to capture long-term dependencies and periodic patterns in time series data \cite{wu2021autoformer}.
\end{quote}

\begin{quote}
DLinear: An MLP-based model \cite{zeng2023transformers} that decomposes the input sequence into trend and seasonal components and applies linear mappings for forecasting.
\end{quote}

\begin{quote}
TimesNet: A CNN-based architecture \cite{wu2022timesnet} that transforms one-dimensional time series into two-dimensional representations through fast Fourier transform (FFT), enabling the modeling of multi-period temporal variations from multiple perspectives.
\end{quote}

\begin{quote}
iTransformer: A Transformer-based time series forecasting architecture \cite{liu2024itransformer} that retains the original Transformer structure but modifies the roles of the attention mechanism and feed-forward network through an inverted data representation.
\end{quote}

\begin{quote}
PatchTST: A Transformer-based framework employing patching and channel independence mechanisms \cite{nie2022time}. It treats local temporal segments as input tokens, enabling effective extraction of long-term dependencies while reducing computational complexity.
\end{quote}

\begin{quote}
ECMWF-HRES: A high-resolution operational numerical weather prediction system developed by the European Centre for Medium-Range Weather Forecasts (ECMWF), which integrates advanced atmospheric dynamical and physical models with data assimilation techniques and is widely recognized as one of the leading global weather forecasting systems \cite{chen2024machine,ling2022multi,owens2018ecmwf}.
\end{quote}

All the experiments are conducted on the same equipment, and the relevant parameters are shown in Supplementary Table 1 and Supplementary Table 2. Among these methods, CNN-LSTM represents a classical deep learning approach for temporal sequence modeling, whereas Autoformer, DLinear, TimesNet, iTransformer and PatchTST are representative state-of-the-art (SOTA) forecasting algorithms widely adopted in recent time series forecasting studies. All data-driven models were trained on the same dataset using an identical batch size of 32 and trained for 20 epochs, ensuring a fair comparison. To further improve the reliability of the evaluation and reduce the influence of randomness, five random seeds (seed = 7, 42, 123, 999, and 2021) were employed for repeated experiments. The parameters for all machine learning models are provided in Supplementary Table 3.

The ECMWF-HRES (European Centre for Medium-Range Weather Forecasts High Resolution system) is one of the most accurate numerical weather prediction models worldwide. During the testing phase, forecasts from ECMWF-HRES were collected for the same locations and periods as the test dataset, covering all selected typhoon events. The forecasts were initialized at 00:00 UTC with matching spatial and hourly temporal resolutions, and the forecast horizons were aligned with those of WDANet. This evaluation strategy ensures consistency and fairness between the proposed model and the operational numerical weather prediction baseline under identical forecasting conditions.

\begin{figure}[h]
\centering
\includegraphics[width=1\textwidth]{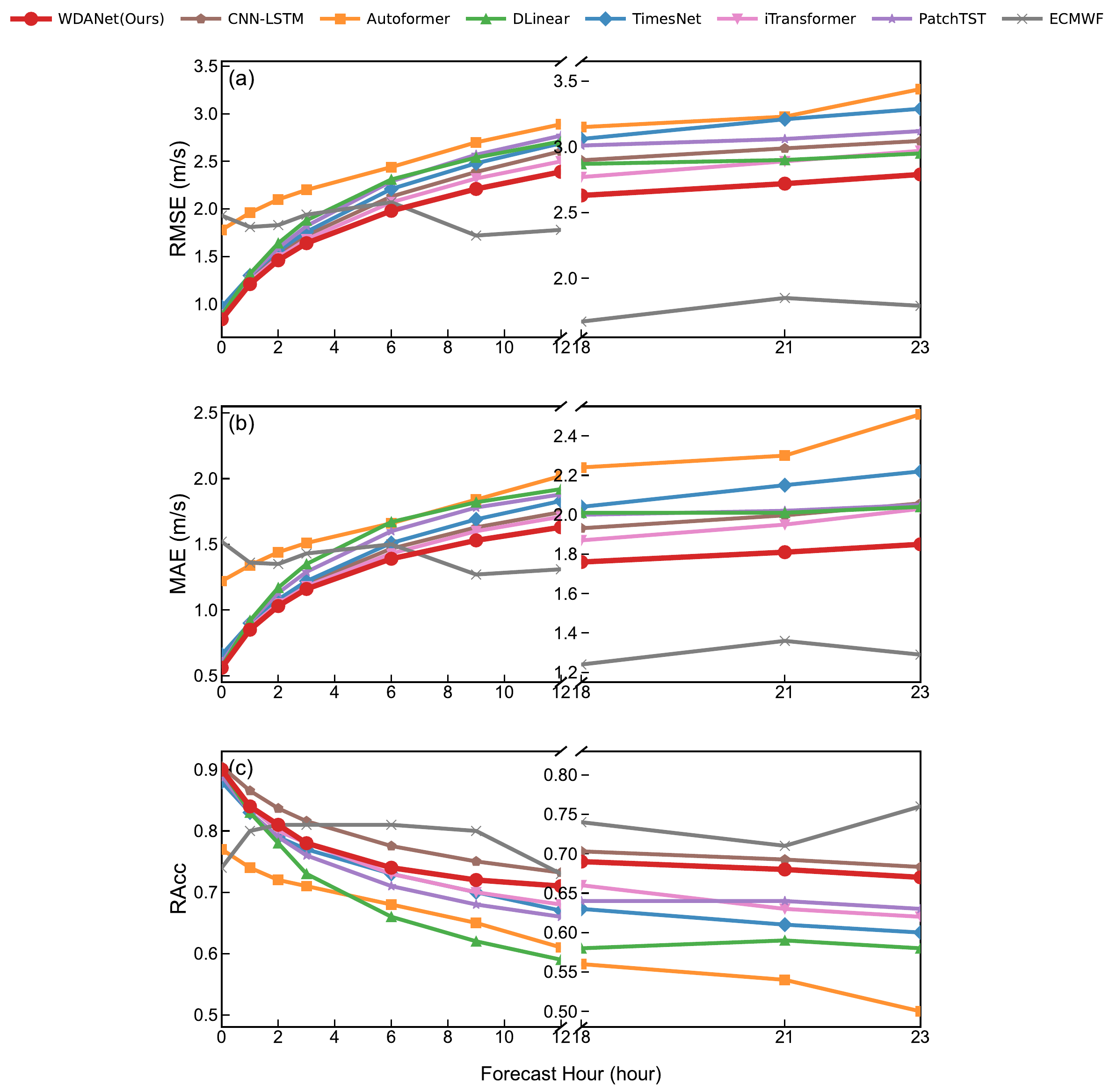}
\caption{\textbf{Performance comparison of different models for 24h wind gust prediction.} 
\textbf{a} is RMSE result. \textbf{b} is MAE result. \textbf{c} is RAcc result. 
Lower RMSE/MAE and higher RAcc indicate better performance. 
A broken x-axis is used to distinguish short-term (0–12 h) and long-term (12–24 h) prediction intervals. 
The proposed model demonstrates competitive and stable performance across different lead times.}
\label{fig:RMSE_MAE_ACC}
\end{figure}

Tables~\ref{tab:rmse}, \ref{tab:mae}, and \ref{tab:racc} present the performance metrics of different models over the 24-hour forecasting horizon, where the RMSE, MAE, and RAcc are all calculated based on hourly forecast results. Figure~\ref{fig:RMSE_MAE_ACC} further illustrates the variation trends of these three metrics with respect to forecast lead time within the 24-hour prediction range. The overall performance metrics are presented in Supplementary Table 4.

As shown in Figure~\ref{fig:RMSE_MAE_ACC}a, the RMSE values of all models exhibit an increasing trend with increasing forecasting lead time, which is mainly attributed to the accumulation of prediction uncertainty and error propagation during wind gust evolution. Compared with the data-driven baseline approaches, the proposed WDANet consistently achieves the lowest RMSE across all forecasting horizons. Specifically, the RMSE values at 1-h and 24-h lead times reach 0.84~m/s and 2.79~m/s, respectively. Compared with the best-performing baseline model iTransformer, WDANet reduces the forecasting error by approximately 5.6\%--6.1\% across different lead times. Compared with Autoformer, the error reduction reaches 18.9\%--52.8\%. 

A similar performance trend is observed for the MAE metric, where WDANet maintains the lowest prediction errors among all data-driven models. At the final forecasting step, WDANet achieves an MAE of only 1.85~m/s, representing improvements of approximately 8.9\% and 26.3\% compared with iTransformer (2.03~m/s) and Autoformer (2.51~m/s), respectively. These results indicate that WDANet provides improved multi-step forecasting capability among data-driven approaches and effectively alleviates error accumulation during extended prediction horizons.

The RAcc results further characterize the relative forecasting capability of different models. CNN-LSTM achieves a slightly higher RAcc than the proposed model, indicating its advantage in preserving the overall temporal variation tendency. This phenomenon may be attributed to the smoothing characteristics of CNN-LSTM, which enable it to capture dominant temporal trends while suppressing certain high-frequency fluctuations. Nevertheless, WDANet still outperforms most other baseline models, with only a marginal decrease of approximately 0.01 compared with CNN-LSTM over the entire forecasting period. Meanwhile, WDANet achieves the lowest RMSE and MAE values, indicating that although its trend consistency is slightly different from CNN-LSTM, it provides more accurate predictions of wind gust intensity.

\begin{figure}[h]
\centering
\includegraphics[width=1\textwidth]{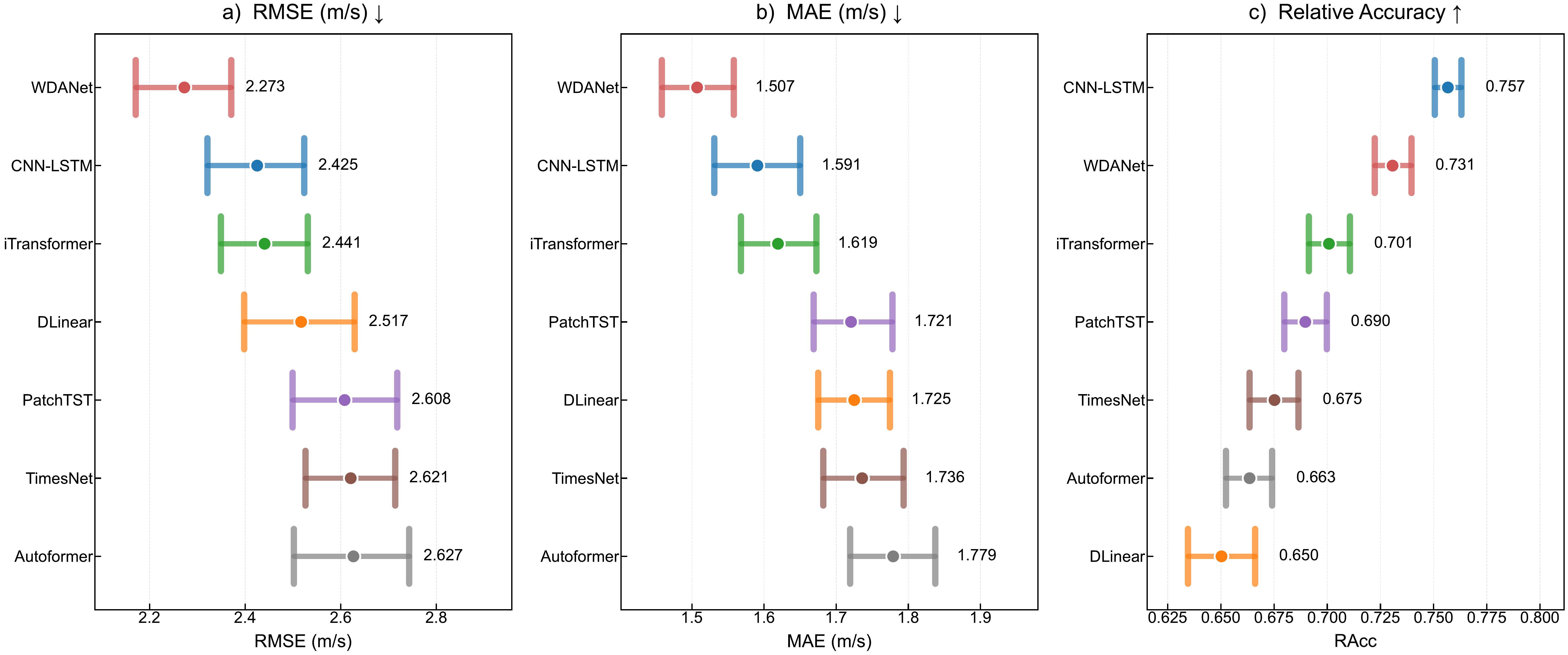}
\caption{\textbf{95\% confidence intervals of model performance obtained from typhoon-event block bootstrap.} 
\textbf{a} is RMSE result. \textbf{b} is MAE result. \textbf{c} is RAcc result. The markers represent the mean values of each metric, and the error bars indicate the 95\% confidence intervals.}
\label{fig:bootstrap}
\end{figure}

To further assess the statistical robustness of model performance, a block bootstrap analysis was conducted based on individual typhoon events rather than hourly samples, using a fixed random seed of 42 with 1,000 resampling iterations. This strategy preserves the temporal dependence within each typhoon process and provides more reliable uncertainty estimation. The 95\% confidence intervals of RMSE, MAE, and RAcc are presented in Figure ~\ref{fig:bootstrap}.

WDANet achieves the lowest RMSE (2.27~m/s) with a narrow confidence interval of [2.17, 2.37]~m/s, which is consistently lower than all baseline models. Similar behavior is observed for MAE, where WDANet obtains the smallest mean value (1.51~m/s) and maintains stable uncertainty across different typhoon events. Although CNN-LSTM achieves a slightly higher RAcc, WDANet provides lower RMSE and MAE, indicating more accurate estimation of wind gust intensity. These results demonstrate that the improvement of WDANet is robust across different typhoon cases rather than being dominated by individual events.

The superior performance of WDANet can be attributed to its frequency-aware forecasting mechanism. By applying stationary wavelet transform, the proposed model decomposes the input sequence into low-frequency and high-frequency components and processes them through independent encoder-decoder pathways. Such a design enables the separation of temporal characteristics at different frequency levels and reduces the interference between components. Compared with treating the input sequence as a unified signal, this frequency decomposition strategy improves the stability of multi-step forecasting. Furthermore, the decoupled architecture may reduce error propagation across forecasting horizons by independently modeling high-frequency variations and low-frequency trends. Consequently, the prediction error increases more slowly over longer forecasting periods, which is consistent with the observed trends in RMSE, MAE, and RAcc.

Compared with ECMWF-HRES, WDANet demonstrates competitive short-term forecasting
capability, particularly within the first 6h. At the 1-h lead time, ECMWF-HRES obtains an RMSE of 1.93~m/s, whereas WDANet achieves an RMSE of 0.84~m/s, corresponding to a reduction of 56.5\%. Similar advantages are observed in MAE and RAcc during the early forecasting period. However, as the forecasting lead time increases, the performance degradation of WDANet becomes more significant than that of ECMWF-HRES, indicating that ECMWF-HRES possesses stronger capability in capturing long-term atmospheric evolution patterns. The comparison with the operational numerical weather prediction system demonstrates that WDANet provides competitive short-term wind gust forecasting performance within 6 h. Combined with its higher computational efficiency, this result provides confidence in the potential application of WDANet in wind power.

\subsection{Ablation Study}\label{sec:ablation}
To further investigate the contribution of individual components, we conducted ablation experiments under five different random seeds to ensure statistical robustness. The results are summarized in Table~\ref{tab:ablation_wavelet} and Table~\ref{tab:ablation_components}, including the effects of wavelet configurations, frequency branches, and attention mechanisms.

The influence of wavelet basis and decomposition depth is reported in Table~\ref{tab:ablation_wavelet}. Under one-level decomposition, sym4 achieved the best performance with an RMSE of 2.28 and MAE of 1.55, while db2 and db4 showed slightly higher errors. Haar obtained relatively worse results, possibly due to its discontinuous basis function, which is less suitable for representing smooth meteorological variations. Increasing the decomposition depth beyond one level did not further improve performance. For sym4, two- and three-level decompositions increased RMSE to 2.59 and 2.51, respectively, indicating that excessive decomposition may reduce the effective temporal information available for sequence modeling.

The role of SWT was examined by comparing the complete model with learnable SWT filters, the fixed-SWT variant, and the model without wavelet decomposition. Removing SWT increased the RMSE from 2.28 to 2.92 and MAE from 1.55 to 1.89, indicating that frequency decomposition provides effective multi-scale representations for wind gust forecasting. Although the RAcc values of the two configurations remain similar, the larger error metrics without SWT demonstrate that wavelet decomposition mainly contributes to reducing prediction deviations. The fixed-SWT variant achieved an RMSE of 2.38, which was better than removing SWT but still inferior to the learnable version. This result suggests that the performance improvement originates not only from frequency decomposition itself but also from the adaptive adjustment of frequency responses.

The contribution of the FiLM-based cross-frequency modulation was further evaluated by removing the FiLM module from the complete architecture. As shown in Table~\ref{tab:ablation_components}, the variant without FiLM achieved an RMSE of 2.30, compared with 2.28 obtained by the full model. Although the MAE value of the w/o FiLM variant was slightly lower, the decrease in RMSE and RAcc indicates that FiLM mainly improves the model's ability to capture large-magnitude forecasting deviations rather than uniformly reducing average errors. This behavior is important for wind gust forecasting under typhoon conditions, where extreme fluctuations contribute disproportionately to prediction uncertainty. By adaptively generating modulation parameters from complementary frequency components, FiLM enables dynamic information exchange between low-frequency background trends and high-frequency transient variations. Without this cross-frequency interaction, the model relies on independently encoded frequency representations, limiting its capability to adjust feature responses according to rapidly changing atmospheric states.

The necessity of the dual-branch architecture was evaluated by retaining only the approximation branch or the detail branch. Both variants showed degraded performance compared with the complete model, with RMSE values of 2.33 and 2.32, respectively. The approximation branch mainly captures low-frequency trends, whereas the detail branch describes short-term fluctuations. The performance gap indicates that accurate wind gust forecasting requires the joint modeling of stable temporal evolution and rapid variations.

The effect of the hybrid attention mechanism was further analyzed by replacing both branches with only global attention or only local attention. The global-attention-only and local-attention-only variants achieved RMSE values of 2.34 and 2.39, respectively, both higher than the complete model. Global attention provides stronger long-range dependency modeling but may weaken the representation of abrupt local changes, while local attention improves short-term feature extraction at the cost of global temporal consistency. The proposed hybrid attention mechanism effectively combines these complementary capabilities, leading to improved forecasting performance.

These results verify that the proposed architecture benefits from the collaborative effects of adaptive SWT, FiLM-based cross-frequency modulation, dual-frequency representations, and hybrid attention. Each component addresses a specific limitation in wind gust forecasting, and their combination enables more accurate and robust prediction under complex typhoon conditions.

\subsection{Performance under extreme typhoon conditions}\label{sec:Performance under extreme typhoon conditions}

In the previous section, the average forecasting performance of different models was evaluated over the complete typhoon-active period. However, extreme wind gust events involve stronger nonlinear variations and larger uncertainties, providing a more challenging scenario for evaluating forecasting robustness. Therefore, based on the same test dataset used in Section~\ref{sec:Comparison methods analysis}, two subsets were further extracted to evaluate model performance under more severe conditions: samples with maximum wind speeds exceeding 15~m/s were classified as typhoon-affected cases, and samples whose maximum wind gust values fell within the top 10\% of all typhoon cases were defined as extreme typhoon cases, with wind gust intensities ranging from 28.52 to 46.85~m/s.

\begin{figure}[h]
\centering
\includegraphics[width=1\textwidth]{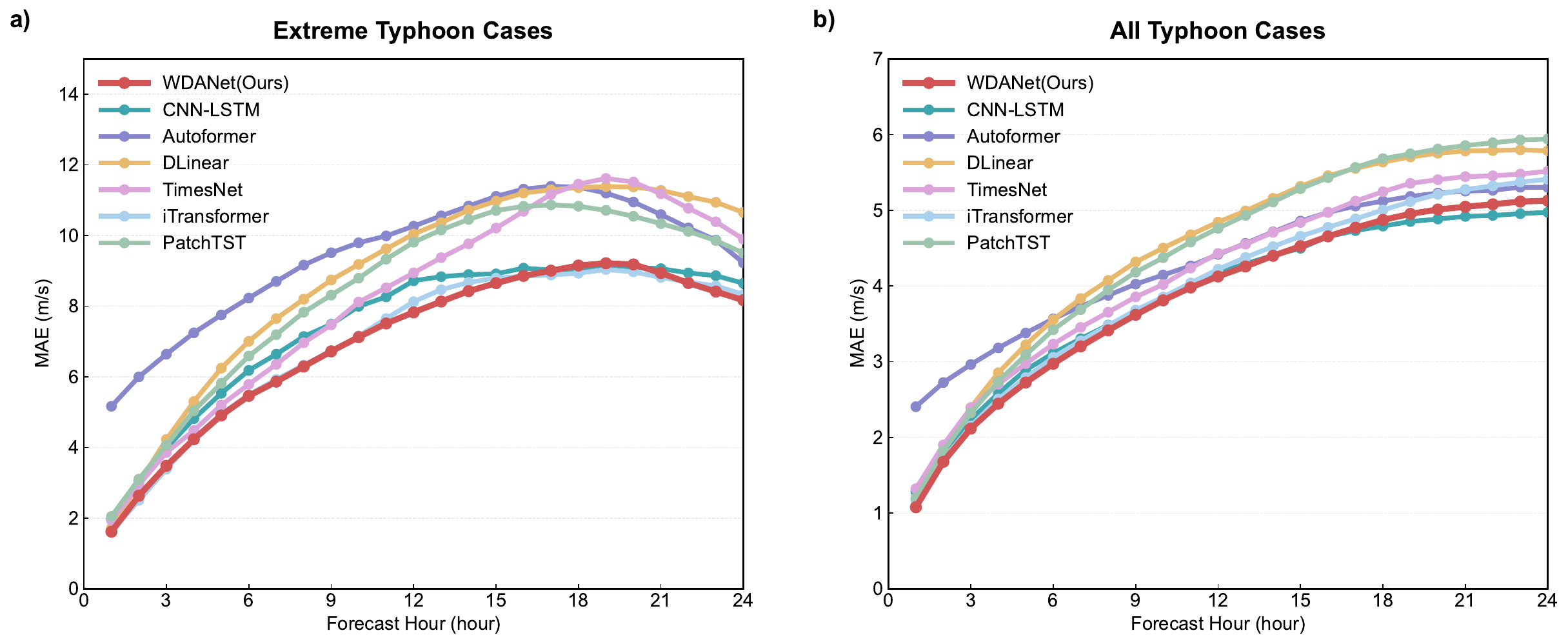}
\caption{\textbf{Performance comparison of multi-step wind gust forecasting under different typhoon scenarios.} 
\textbf{a} MAE under typhoon conditions with maximum wind gusts ranging from 28.52 m/s to 46.85 m/s. \textbf{b} MAE under typhoon conditions with maximum wind gusts exceeding 15 m/s. The horizontal axis denotes the forecast horizon, and the vertical axis represents the error.}
\label{fig:comparison_extreme_and_all}
\end{figure}

Figure~\ref{fig:comparison_extreme_and_all}a presents the hourly MAE of all models over the 24-h forecasting horizon under extreme typhoon cases. With increasing forecasting lead time, all models exhibit a rapid increase in MAE, indicating that the strong nonlinear characteristics and increased uncertainty of extreme typhoon events significantly increase the difficulty of forecasting. During the early forecasting period, iTransformer achieves relatively better performance. However, after 12 h, WDANet consistently maintains the lowest MAE, demonstrating its superior robustness under extreme wind gust conditions. 

Under extreme typhoon conditions, WDANet achieves the lowest 24-h averaged MAE of 7.02~m/s, outperforming CNN-LSTM (7.47~m/s), iTransformer (7.10~m/s), PatchTST (8.21~m/s), TimesNet (8.28~m/s), DLinear (8.90~m/s), and Autoformer (9.46~m/s). Compared with Autoformer, WDANet reduces the error by 25.8\%, while the improvements over TimesNet and DLinear reach 15.2\% and 21.1\%, respectively. This demonstrates that WDANet is more capable of estimating extreme wind gust intensity under highly variable typhoon conditions. Averaged over the entire 24-h forecasting horizon, WDANet also achieves a competitive overall MAE of 3.87 m/s for all typhoon-affected cases, confirming its applicability across different wind intensity regimes.

Beyond the overall average, the hour-by-hour MAE variation for all typhoon-affected cases is further examined (Figure~\ref{fig:comparison_extreme_and_all}b). WDANet achieves the lowest MAE over most forecasting horizons, while CNN-LSTM shows slightly lower errors during the final few hours. This indicates that CNN-LSTM can effectively capture dominant temporal trends under moderate conditions, which is consistent with our observations in the overall performance comparison. In contrast, WDANet provides stronger advantages for rapidly varying and high-amplitude wind gust events. The standard deviation analysis in Supplementary Fig.~1 and Fig.~2 further shows that WDANet maintains relatively lower prediction variability across forecasting horizons, demonstrating improved consistency under diverse typhoon scenarios.

Overall, these results highlight the effectiveness of WDANet for extreme wind gust forecasting. By combining adaptive frequency decomposition and cross-frequency feature interaction, WDANet reduces error accumulation and provides reliable predictions under nonlinear and rapidly evolving typhoon conditions.

\subsection{Case study of Typhoon Yagi}\label{sec:Case study of Typhoon Yagi}

\begin{figure}[h]
\centering
\includegraphics[width=0.90\textwidth]{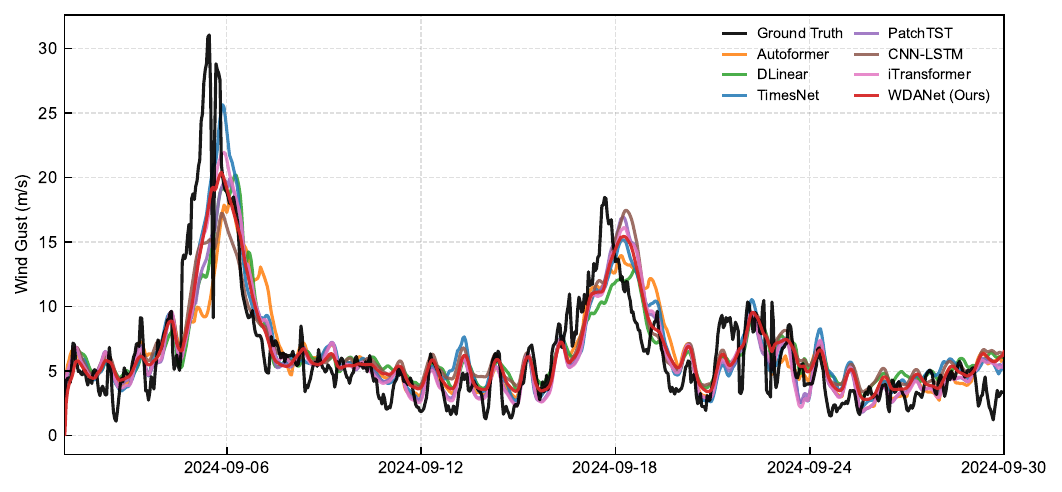}
\caption{\textbf{Case study of Typhoon Yagi.} 
Comparison between predicted results and ground truth in Haikou, Hainan Province, China (20.0° N, 110.25° E) under the input-24/predict-24 setting during Typhoon Yagi, from September 2 to September 30, 2024.}
\label{fig:Yagi_prediction_All_Models}
\end{figure}

To evaluate the practical forecasting capability of WDANet under extreme typhoon conditions, Typhoon Yagi (2411) is selected as a representative case study. As the strongest autumn typhoon to make landfall in China since 1949, Typhoon Yagi caused the collapse of multiple wind turbines in Hainan Province during its passage \cite{dialogue2024yagi}, resulting in considerable economic losses and highlighting the importance of reliable gust forecasting for the safe operation of wind energy systems. Haikou, Hainan Province, China (20.0$^\circ$N, 110.25$^\circ$E), is selected as the target location. During this event, the wind field exhibits pronounced non-stationary characteristics, with irregular fluctuations before landfall, rapid intensification during the main impact period, and subsequent decay, resulting in substantial variations in gust intensity (Figure~\ref{fig:Yagi_prediction_All_Models}). Based on the experimental settings described above, model performance is evaluated from two complementary perspectives: forecasting accuracy over the entire month containing Typhoon Yagi and predictive capability during the extreme high-wind period.

Supplementary Fig. 3 summarizes the forecasting errors of different models over the entire monthly period associated with Typhoon Yagi. This monthly-scale evaluation reflects the capability of different models to continuously track both background wind variations and typhoon-induced disturbances. WDANet achieves the best overall performance among all compared models, with an MAE of 1.70~m/s and an RMSE of 2.50~m/s. Compared with the second-best model, iTransformer (MAE: 1.73~m/s, RMSE: 2.57~m/s), WDANet reduces the MAE and RMSE by approximately 1.7\% and 2.7\%, respectively. Compared with TimesNet (MAE: 1.82~m/s, RMSE: 2.61~m/s), WDANet further reduces the MAE and RMSE by approximately 6.6\% and 4.2\%, respectively. These results demonstrate that WDANet can effectively capture the complex temporal variations associated with typhoon-induced gusts and maintain stable forecasting performance over extended prediction periods.

\begin{figure}[h]
\centering
\includegraphics[width=1\textwidth]{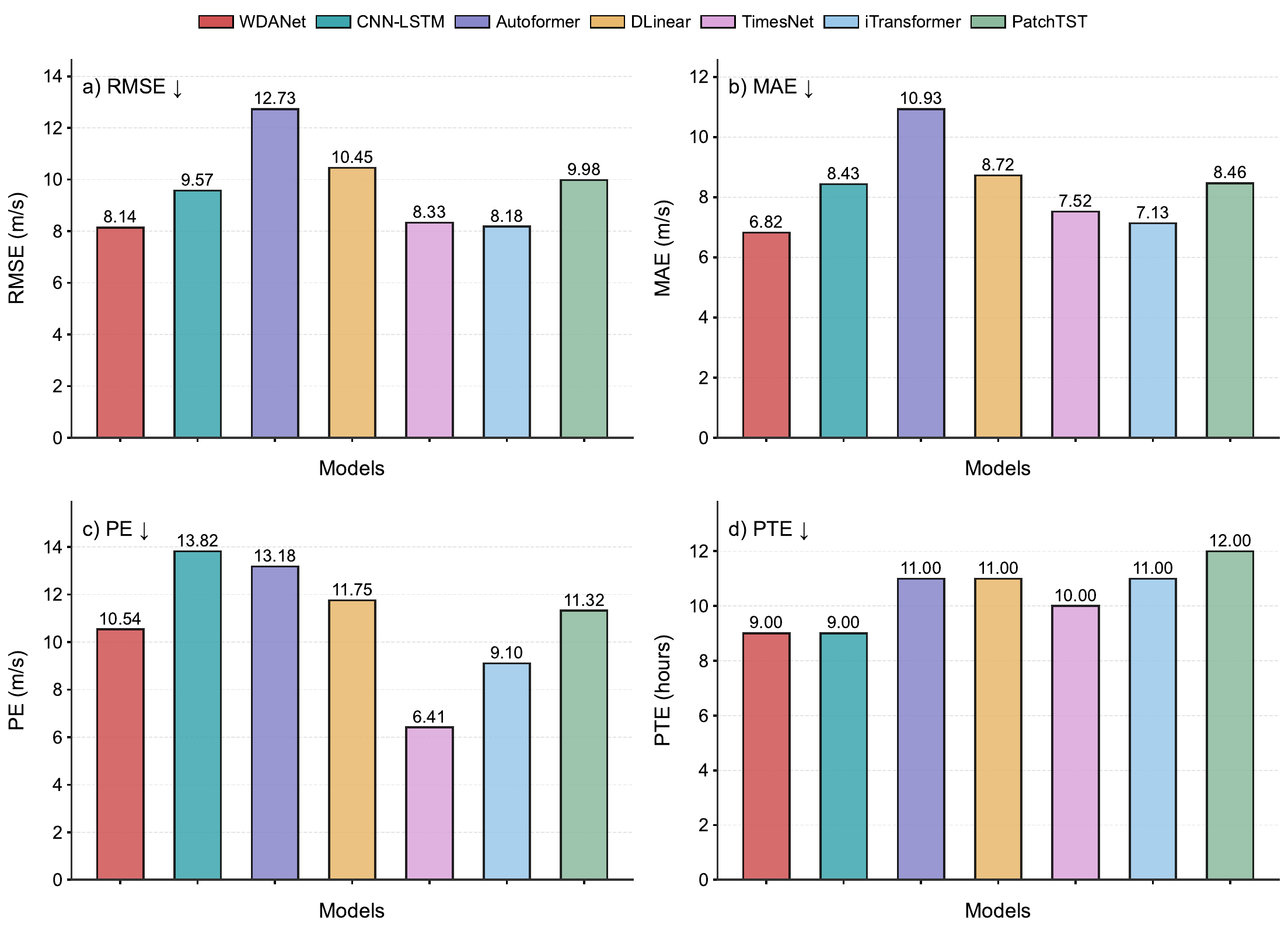}
\caption{\textbf{Forecasting performance of different models during the most intense 24-h gust period of Typhoon Yagi.} 
Performance comparison of different forecasting models during the most intense 24-h gust period of Typhoon Yagi. The selected period represents the consecutive 24 hours with the highest average gust intensity, covering the most challenging extreme wind conditions. The evaluation includes \textbf{a} RMSE, \textbf{b} MAE, \textbf{c} peak error (PE), and \textbf{d} peak time error (PTE). Lower values indicate better performance.}
\label{fig:Yagi_extreme_MAE_RMSE_PE_PTE}
\end{figure}

To further investigate the model capability under extreme wind conditions, Figure~\ref{fig:Yagi_extreme_MAE_RMSE_PE_PTE} presents the forecasting results during the most intense 24-hour high-wind period of Typhoon Yagi. This period covers the rapid strengthening and decay stages of the typhoon, with peak gust intensity exceeding 31~m/s, representing a highly challenging scenario characterized by strong nonlinear fluctuations. WDANet achieves the lowest MAE of 6.82~m/s and the lowest RMSE of 8.14~m/s among all compared models. Compared with TimesNet, WDANet reduces the MAE by approximately 9.3\% and achieves comparable RMSE performance (8.14~m/s versus 8.33~m/s). Compared with Autoformer, WDANet reduces the MAE and RMSE by approximately 37.6\% and 36.0\%, respectively. These results indicate that WDANet exhibits stronger robustness in responding to rapidly varying extreme gust conditions.

In addition to conventional error metrics, peak-related metrics are employed to further evaluate the capability of different models in characterizing extreme gust events. As shown in Figure~\ref{fig:Yagi_extreme_MAE_RMSE_PE_PTE}, WDANet achieves the lowest peak time error (PTE) of 9 hours, indicating superior capability in identifying the occurrence timing of extreme gusts. The accurate peak timing prediction highlights the timeliness of WDANet in capturing the onset of hazardous wind conditions, while its lower sequence-level errors demonstrate reliable forecasting performance under rapidly evolving typhoon environments. Although TimesNet achieves a smaller peak error (PE), indicating better reconstruction of the maximum gust magnitude at a specific time point, its higher MAE suggests less consistent prediction performance throughout the entire extreme period. Therefore, WDANet provides a better balance between overall forecasting accuracy, temporal response capability, and extreme-event robustness.

The different rankings among MAE, RMSE, PE, and PTE reveal the complementary characteristics of these evaluation metrics. MAE and RMSE quantify the overall deviation across the forecasting sequence, whereas PE focuses only on the error at the maximum gust point and may be sensitive to isolated peak reconstruction. Therefore, a smaller PE does not necessarily indicate superior extreme-event forecasting capability. For Typhoon Yagi, TimesNet obtains the lowest PE, suggesting a more accurate estimation of the maximum gust amplitude, while WDANet achieves lower sequence-level errors and more accurate peak timing. This advantage can be attributed to the dual-frequency decomposition and adaptive fusion strategy of WDANet, which enables simultaneous modeling of slowly varying wind evolution and high-frequency gust fluctuations. Consequently, WDANet provides more stable, timely, and reliable forecasts for extreme typhoon-induced gust events, demonstrating its potential for wind energy safety management and early-warning applications.

\section{Discussion}\label{sec:Discussion}

Accurate and reliable forecasting of extreme typhoon events is of great importance for offshore wind power development, particularly in regions frequently affected by severe weather conditions. In this study, we propose WDANet, an effective multivariate data-driven framework specifically designed for wind gust forecasting (WGF). Compared with existing deep learning models, WDANet achieves superior performance in terms of RMSE, MAE, and RAcc. Compared with the ECMWF-HRES numerical weather prediction system, WDANet demonstrates significant advantages in short-term forecasting, achieving reductions of more than 50\% in RMSE and MAE under certain forecasting horizons. Extensive evaluations on typhoon cases, extreme typhoon events, and the representative Typhoon Yagi case demonstrate that WDANet can provide more accurate and timely wind gust forecasts for key offshore wind energy regions.

In short-term forecasting, WDANet directly learns the nonlinear mapping relationship between historical wind gust data and future wind gust predictions, enabling superior predictive capability. In contrast, ECMWF-HRES relies on physical process parameterizations to represent atmospheric evolution, where parameterization schemes inherently introduce approximation errors \cite{brenowitz2018prognostic}, and uncertainties in the initial conditions are unavoidable. However, for longer forecasting horizons, ECMWF-HRES benefits from the physical constraints imposed by atmospheric governing equations and exhibits stronger long-term stability. Since WDANet does not incorporate explicit physical constraints, forecasting errors may be nonlinearly amplified during autoregressive prediction.

Despite these advantages, several limitations remain in this study. The experiments are conducted using ERA5 reanalysis data with a spatial resolution of $0.25^{\circ}\times0.25^{\circ}$. Although this resolution is widely adopted in atmospheric studies and provides relatively high-quality meteorological information, it may still be insufficient to capture finer-scale wind variations. It should be noted that the release of ERA5 data is subject to significant latency. According to the official documentation provided by the Copernicus Climate Change Service, ERA5 data are typically available with a delay of approximately 5 days. Therefore, although this study demonstrates the technical feasibility and accuracy of grid-point wind gust forecasting based on historical reanalysis data, the proposed model cannot directly support real-time operational deployment. In practical operational scenarios, the model's inputs must be replaced with near-real-time data streams. Future research should focus on evaluating the model's performance after integrating real-time forecast inputs, in order to validate its true operational potential. Furthermore, spatial information is not incorporated into the current framework to avoid potential deployment challenges associated with cross-station data sharing and transmission constraints. The evaluation is currently limited to typhoon events in the western North Pacific, and the applicability of WDANet to other ocean basins requires further investigation. Furthermore, WDANet still exhibits limitations in reproducing extremely high wind gust peaks. Similar behavior is also observed in other deep learning models evaluated in this study, suggesting that accurate extreme peak reconstruction remains a challenging problem for data-driven approaches. This limitation may be partly associated with the limited availability of extreme wind events in the training dataset, which restricts the model's ability to learn rare high-intensity gust patterns.

Future studies can be extended in several directions. First, numerical weather prediction outputs, such as ECMWF forecasts, can be incorporated as auxiliary inputs to combine physical constraints with data-driven representations, which may improve long-term forecasting and extreme peak prediction. Second, transfer learning strategies can be explored to adapt the proposed framework to different geographical regions and offshore wind farm locations. Third, data augmentation and cost-sensitive learning approaches can be investigated to alleviate the impact of the limited availability of extreme event samples during training. As wind power generation continues to expand, public concern regarding energy stability and reliability has intensified. Accurate wind gust forecasts can strengthen public confidence in wind energy as a sustainable power source, positioning it as a more mature and competitive energy option.

\section{Methods}\label{sec:Methods}

\subsection{Dataset and preprocessing}\label{sec:Dataset and preprocessing}

Meteorological data used in this study are derived from the ERA5 reanalysis dataset provided by the European Centre for Medium-Range Weather Forecasts (ECMWF) \cite{era5_single_levels_2023}. The spatial resolution of the selected data is 0.25° × 0.25° and the temporal resolution is 1 h. The selected variables include sea level pressure, surface pressure, 2 m air temperature, surface temperature, dew point temperature, relative humidity, and 10 m wind gust since previous post-processing. Relative humidity is computed as the ratio of actual vapor pressure to saturation vapor pressure. The pressure field of a tropical cyclone plays a crucial role in determining its wind structure, as horizontal pressure gradients govern the intensity and organization of wind fields \cite{he2019toward}. Meanwhile, temperature and moisture jointly influence the TC pressure field through their effects on air density\cite{snaiki2017modeling}. These factors collectively affect the vertical distribution of pressure deficit and hence the resulting wind field within the TC boundary layer, and they also jointly influence the development of turbulence and the efficiency of vertical momentum transport within the boundary layer, which are the primary mechanisms responsible for the generation and amplification of wind gusts.

A total of 400 typhoon events affecting the study region during the period from 1960 to 2025 are identified. For each typhoon event, hourly ERA5 reference gust data at the grid point corresponding to the station are extracted for the entire month during which the typhoon occurs. This strategy ensures that each sample contains a continuous time series covering pre-typhoon, typhoon, and post-typhoon phases, rather than isolated extreme segments. As a result, the dataset captures both stable background conditions and highly non-stationary dynamics within a unified temporal context, allowing the model to learn transitions between normal and extreme regimes.

\begin{figure}[h]
\centering
\includegraphics[width=0.80\textwidth]{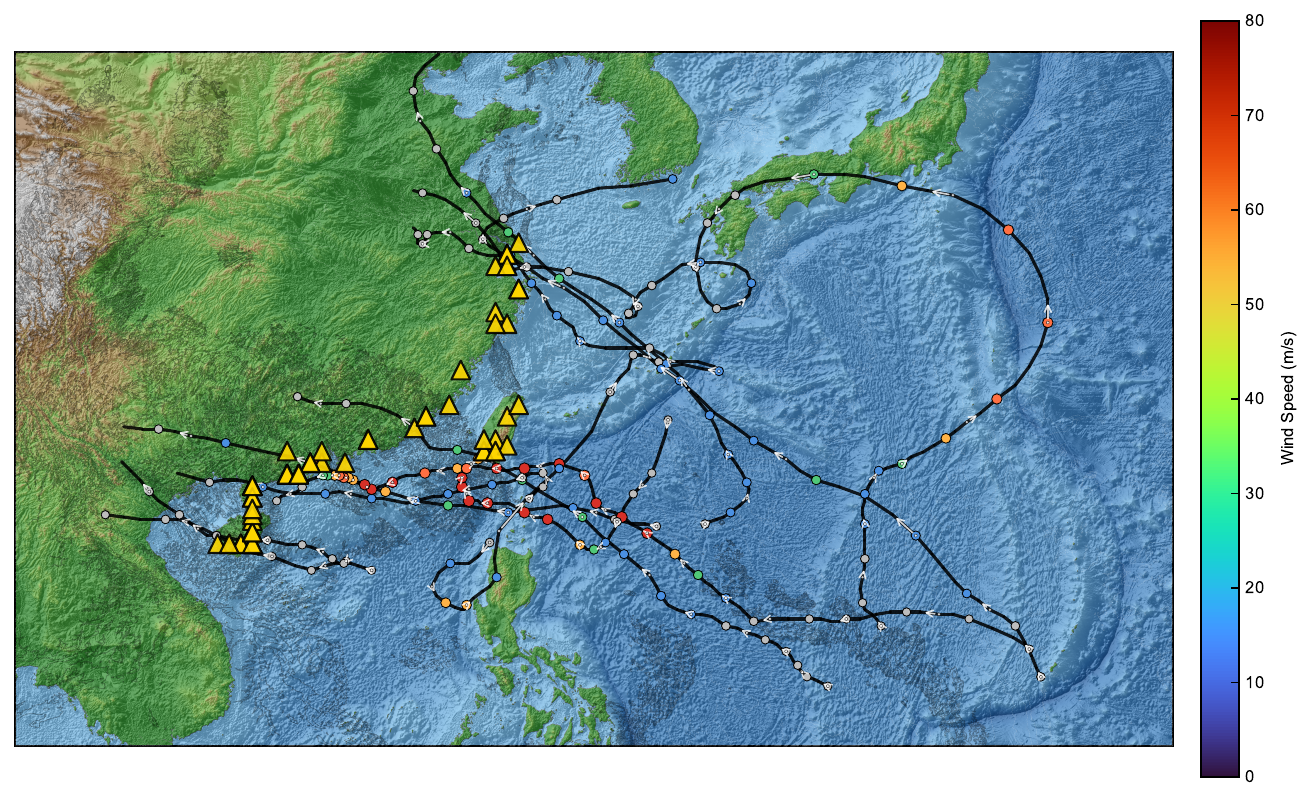}
\caption{\textbf{Western North Pacific Typhoon Tracks (100°E 150°E, 10°N 40°N.)
Wind-Speed Gradient and Intensity Classification} 
Spatial distribution of typhoon tracks over the western North Pacific basin (100°E–150°E, 10°N–40°N). Tracks are color-coded according to tropical cyclone intensity classification, with the wind-speed gradient (units: m/s) indicated by the vertical color bar on the right. Yellow triangles indicate the ERA5 reanalysis data extraction locations.}
\label{fig:Typhoon_Tracks}
\end{figure}

All samples were organized in chronological order to preserve the temporal structure of the data. Missing values were first handled using forward filling, while missing values at the beginning of sequences were filled using backward filling. Any remaining missing values were replaced with zeros. To determine an appropriate historical input length, the temporal dependency of wind gusts was investigated using autocorrelation analysis (Supplementary Fig. 4). The results show that gust autocorrelation remains significant within the first 24 h but decreases rapidly at longer time lags. Therefore, all models adopted a 24-hour sliding window to construct input sequences, with a sliding step of 1 hour. All variables were normalized using min-max normalization to ensure consistent feature scales and improve training stability. All preprocessing parameters, including statistics for missing-value imputation and normalization parameters, were calculated exclusively from the training set. The validation and test sets were transformed using the parameters derived from the training set to prevent any form of information leakage. The dataset was split chronologically into training (70\%, including 280 typhoon events), validation (15\%, including 60 typhoon events), and test (15\%, including 60 typhoon events) sets. The split was performed at the file level to ensure that continuous time-series samples from the same file were not distributed across different subsets, thereby enabling a reliable evaluation of the model's generalization capability. The spatial distribution of the 60 selected typhoon events in the test set and the corresponding ERA5 extraction locations is illustrated in Figure~\ref{fig:Typhoon_Tracks}. The dataset covers the western North Pacific region (100°E–150°E, 10°N–40°N), where tropical cyclones frequently occur and significantly affect offshore wind energy regions.

\subsection{Task formulation}\label{sec:Task formulation}

The wind gust forecasting (WGF) task under typhoon conditions is formulated as a multi-step time series forecasting problem. Given a sequence of historical meteorological reanalysis data over the past $M$ time steps, the objective is to predict the future wind gust evolution over a forecasting horizon of length $H$.

At each time step $i$, the atmospheric state is represented as a vector:
\begin{equation}
\mathbf{x}_{i} = 
\left[
\mathrm{Slp}_{i}, \mathrm{Sp}_{i}, \mathrm{T2m}_{i}, \mathrm{St}_{i}, \mathrm{Td}_{i}, \mathrm{Rh}_{i}, G_{i}
\right]^{\mathrm{T}}
\in \mathbb{R}^{d},
\end{equation}
where $\mathrm{Slp}$, $\mathrm{Sp}$, $\mathrm{T2m}$, $\mathrm{St}$, $\mathrm{Td}$, and $\mathrm{Rh}$ denote sea level pressure, surface pressure, 2-meter air temperature, surface temperature, dew point temperature, and relative humidity, respectively, while $G_{i}$ represents the 10m wind gust since previous post-processing at time step $i$.

Given the historical sequence:
\begin{equation}
\mathbf{x}_{N-M:N-1} =
\left[
\mathbf{x}_{N-M}, \mathbf{x}_{N-M+1}, \ldots, \mathbf{x}_{N-1}
\right],
\end{equation}

the goal is to learn a nonlinear mapping:
\begin{equation}
\hat{\mathbf{G}}_{N:N+H-1} = f\left(\mathbf{x}_{N-M:N-1}\right),
\end{equation}
where $\hat{\mathbf{G}}_{N:N+H-1} \in \mathbb{R}^{H}$ denotes the predicted wind gust sequence over the next $H$ time steps.

\subsection{Learnable stationary wavelet transform module}\label{sec:Learnable stationary wavelet transform module}

To capture the frequency-dependent characteristics of wind gust signals, we employ the Stationary Wavelet Transform (SWT) \cite{nason1995stationary} to decompose the input multivariate sequence into frequency-specific components. The implementation follows the convolution-based formulation described in \cite{chen2025simpletm}. Unlike the conventional Discrete Wavelet Transform (DWT), SWT eliminates the downsampling operation at each decomposition level, thereby preserving all temporal samples and achieving shift invariance. This property avoids the shift sensitivity introduced by decimation in DWT, where small temporal displacements may lead to significant variations in wavelet coefficients \cite{coifman1995translation,nason1995stationary}. Consequently, SWT maintains the temporal correspondence between slowly varying background components and transient fluctuation components, which is particularly important for accurately localizing short-duration wind gust events.

Specifically, the SWT is implemented using one-dimensional convolutions with a pair of low-pass and high-pass filters, denoted as $h$ and $g$. Following the \emph{\`a trous} algorithm, the filters are progressively dilated at different decomposition levels. Let $\mathbf{a}^{(0)}_t=\mathbf{x}_t$ denote the input sequence. At decomposition level $s$, the approximation and detail coefficients are obtained as

\begin{equation}
\mathbf{a}^{(s+1)}_t = \sum_{k} h^{(s)}(k)\mathbf{a}^{(s)}_{t+k}, \quad
\mathbf{d}^{(s+1)}_t = \sum_{k} g^{(s)}(k)\mathbf{a}^{(s)}_{t+k},
\end{equation}

where $h^{(s)}$ and $g^{(s)}$ denote the dilated filters obtained by inserting $2^{s}-1$ zeros between adjacent coefficients of the original filters. The filtering operation is implemented as one-dimensional convolution (equivalent to cross-correlation in deep learning frameworks), which preserves the sequence length and maintains temporal alignment between different frequency components.

Rather than directly adopting fixed wavelet filters, the proposed method initializes the low-pass and high-pass kernels using the Symlet-4 (sym4) wavelet basis and allows them to be optimized during model training. The sym4 initialization provides a physically meaningful frequency decomposition prior, while the subsequent optimization enables the filters to adapt their spectral selectivity according to the intrinsic characteristics of wind gust signals. As illustrated in Supplementary Fig. 5, the learned filters preserve the fundamental low-pass and high-pass characteristics of the initialized wavelet basis while adjusting their frequency responses and spectral weighting. This adaptive refinement allows the decomposition to better represent the distinct temporal-frequency behaviors of wind gust variations, where large-scale atmospheric evolution and short-lived gust fluctuations exhibit different characteristic variability patterns.

In this work, a single-level SWT decomposition is employed, producing a low-frequency component $\mathbf{A}=\mathbf{a}^{(1)}$ and a high-frequency component $\mathbf{D}=\mathbf{d}^{(1)}$. Unlike \cite{chen2025simpletm}, where SWT is mainly used for multi-scale token generation, the decomposed components in this study are directly fed into two dedicated forecasting branches. This frequency-aware architecture enables different branches to specialize in modeling slowly varying atmospheric evolution and rapidly changing gust fluctuations.

\begin{figure}[h]
\centering
\includegraphics[width=0.90\textwidth]{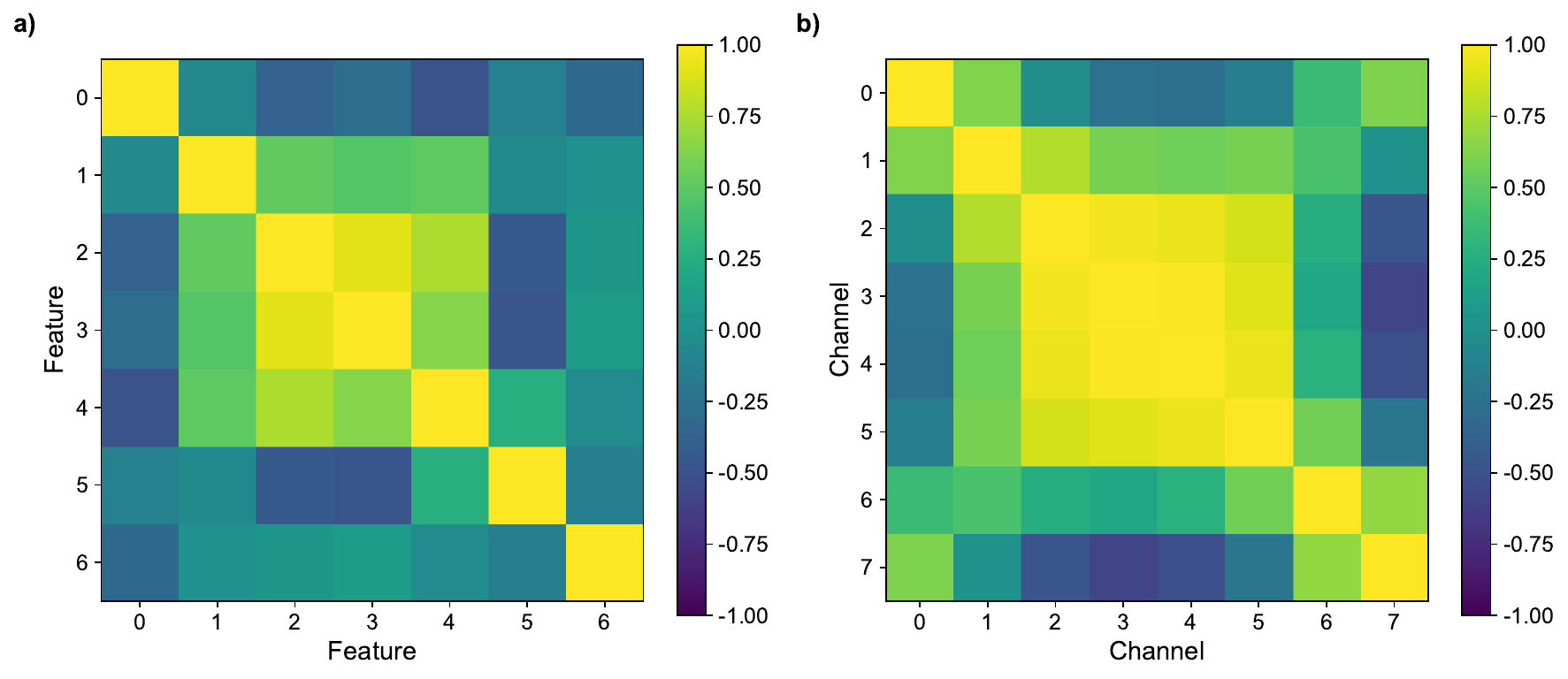}
\caption{\textbf{Correlation analysis of input features and learned representations based on the Symlet-4 wavelet decomposition.} 
\textbf{a} Correlation matrix of the original features. \textbf{b} Correlation matrix of the learned channel representations after wavelet-based decomposition.}
\label{fig:Correlation of fliters}
\end{figure}

To further investigate the impact of the learned wavelet decomposition on feature organization, the correlation matrices before and after SWT processing are visualized in Figure~\ref{fig:Correlation of fliters}. The original meteorological variables exhibit relatively weak and diffuse correlation patterns (Figure~\ref{fig:Correlation of fliters}a), reflecting heterogeneous relationships among different atmospheric variables. In contrast, the representations obtained after the sym4-initialized adaptive SWT exhibit more structured block-like correlation patterns (Figure~\ref{fig:Correlation of fliters}b). This observation suggests that the proposed decomposition facilitates the organization of frequency-dependent meteorological information while preserving the intrinsic relationships among input variables. 

The selection of a single decomposition level is empirically validated through an ablation study (see Section~\ref{sec:ablation}). The results demonstrate that $m=1$ consistently achieves the best forecasting performance. Increasing the decomposition depth does not provide additional improvements and may even degrade performance due to excessive frequency separation, information fragmentation, and the amplification of noise components in higher-frequency bands. Overall, the proposed sym4-initialized adaptive SWT module provides a physically motivated frequency decomposition mechanism that separates slowly varying atmospheric evolution from transient gust fluctuations, enabling subsequent forecasting components to effectively capture the multi-timescale characteristics of wind gust signals.

\subsection{Frequency-aware encoder-decoder architecture}\label{sec:encoder-decoder}
\begin{figure}[h]
\centering
\includegraphics[trim = 0 3cm 0 3cm, clip,width=0.90\textwidth]{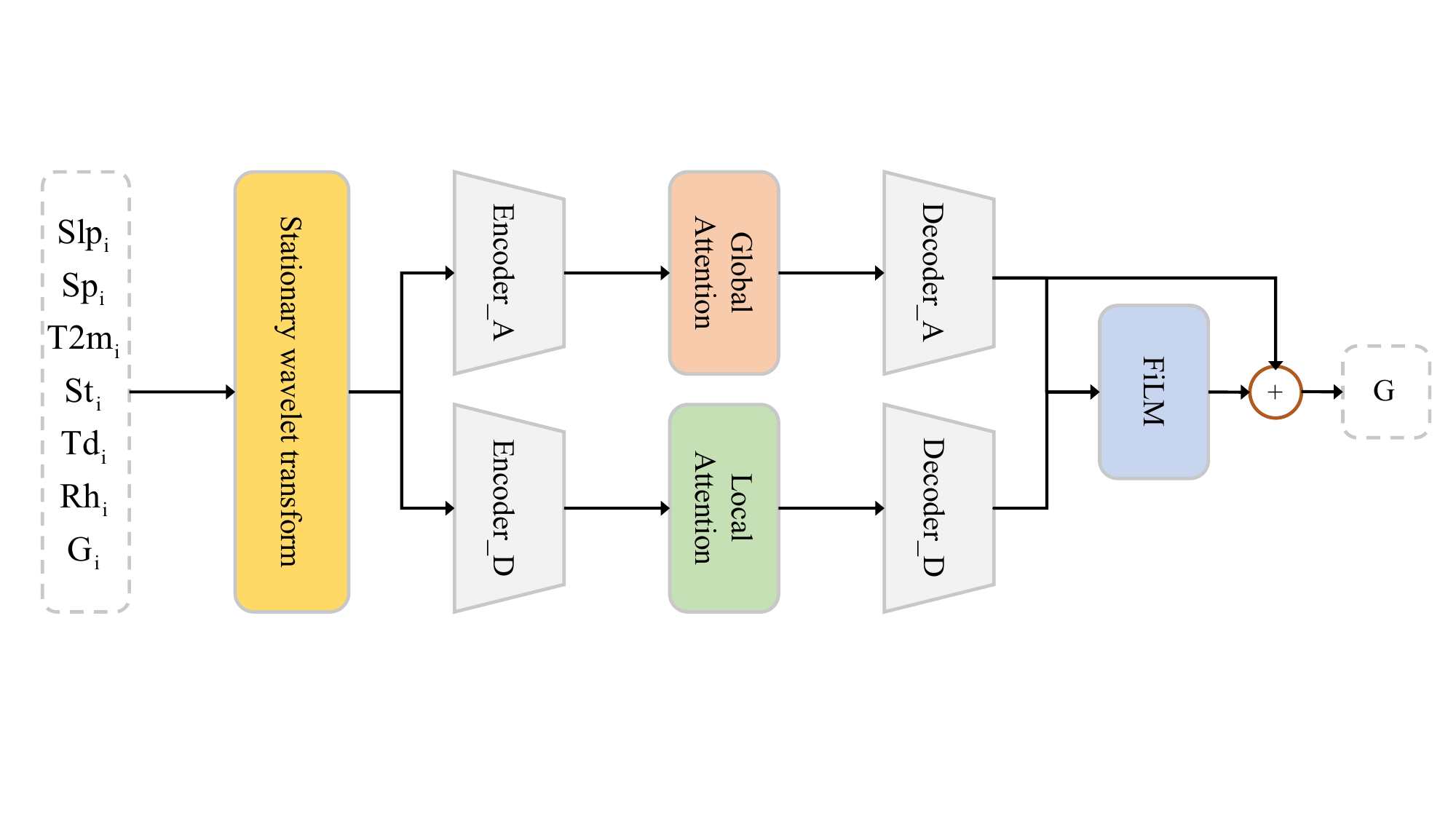}
\caption{\textbf{Architecture of WDANet:} an illustration of the dual-branch network with SWT decomposition, separate encoder-decoders (global/local attention), and FiLM-based cross-frequency modulation for final prediction.}
\label{fig:architecture}
\end{figure}

The overall architecture of WDANet is illustrated in Figure~\ref{fig:architecture}. 
After frequency decomposition by SWT, the input meteorological sequence is separated into
a low-frequency component $\mathbf{A}$ and a high-frequency component $\mathbf{D}$.
These two components are processed by two independent encoder-decoder branches to capture
different temporal characteristics. Specifically, the low-frequency branch focuses on
large-scale atmospheric evolution, while the high-frequency branch models transient wind
gust fluctuations. To establish interactions between the two frequency components, a
FiLM-based cross-frequency modulation module is introduced, where the low-frequency
representation adaptively recalibrates the high-frequency features. Finally, the outputs
from the two branches are reconstructed through additive fusion to obtain the final wind gust prediction.
\subsubsection{Encoder}\label{sec:Encoder}

While the SWT-based decomposition captures multi-scale frequency characteristics of meteorological variables, it does not explicitly model temporal dependencies across time steps. To address this limitation, two structurally identical but parameter-independent bidirectional LSTM encoders are employed to independently process the trend component $\mathbf{A}$ and the fluctuation component $\mathbf{D}$.

For each branch, let the input sequence be denoted as $\mathbf{X} \in \mathbb{R}^{L \times C}$, where $L$ is the sequence length and $C$ is the number of variables. The sequence is first projected into a latent space through a linear embedding layer:
\begin{equation}
\mathbf{Z} = \mathbf{X} \mathbf{W}_{\text{emb}},
\end{equation}
where $\mathbf{W}_{\text{emb}} \in \mathbb{R}^{C \times D}$ is the learnable projection matrix and $D$ is the embedding dimension.

The embedded sequence is then processed by the BiLSTM:
\begin{equation}
\mathbf{H}, (\mathbf{h}_e, \mathbf{c}_e) = \text{BiLSTM}(\mathbf{Z}),
\end{equation}
where $\mathbf{H} \in \mathbb{R}^{L \times 2H}$ denotes the sequence of hidden states, and $\mathbf{h}_e, \mathbf{c}_e \in \mathbb{R}^{S_1 \times 2H}$ are the final hidden and cell states, with $S_1$ being the number of LSTM layers and $H$ the hidden dimension of each direction.

The two encoders share an identical structure but do not share parameters, allowing each branch to specialize in modeling distinct temporal dynamics associated with different frequency bands. The encoded representations $\mathbf{H}$ are subsequently passed to the decoder through attention mechanisms for generating wind gust predictions.

\subsubsection{Attention mechanisms}\label{sec:Attention mechanisms}

To effectively utilize the encoded temporal representations, two distinct attention mechanisms are designed for the trend and fluctuation branches, respectively. Both mechanisms compute a set of attention weights over the encoder outputs, which are subsequently used by the decoder to construct context vectors.

For the trend branch processing $\mathbf{A}$, global attention is adopted to capture long-range dependencies. The attention weights are computed as
\begin{equation}
\alpha_{t,i}^{(\mathbf{A})} = \text{softmax}\Big(\mathbf{v}^{\top}\tanh\big(\mathbf{W}[\mathbf{s}_{t-1}^{(\mathbf{A})}; \mathbf{H}_i^{(\mathbf{A})}]\big)\Big),
\end{equation}
where $\mathbf{s}_{t-1}^{(\mathbf{A})}$ denotes the decoder hidden state at the previous time step, $\mathbf{H}_i^{(\mathbf{A})}$ is the $i$-th encoder output, $\mathbf{W}$ and $\mathbf{v}$ are learnable parameters, and the softmax is taken over all encoder time steps $i=1,\dots,L$. This global formulation allows the model to attend to any position in the input sequence when generating trend predictions.

To better capture short-term and high-frequency fluctuations, a local attention mechanism with a learnable temporal prior is introduced in the fluctuation branch. Unlike global attention, which considers all time steps equally, this mechanism restricts the attention focus to a dynamically predicted local region, thereby improving sensitivity to transient variations.

Specifically, a soft alignment position is first predicted from the decoder hidden state:
\begin{equation}
p_t = L \cdot \sigma(\mathbf{W}_p \mathbf{s}_{t-1}^{(\mathbf{D})}),
\end{equation}
where $L$ denotes the sequence length and $\sigma(\cdot)$ is the sigmoid function that ensures the predicted position lies within a valid temporal range.

Next, a content-based compatibility score between the decoder state and encoder outputs is computed using a dot-product form:
\begin{equation}
e_{t,i} = \langle \mathbf{W}\mathbf{s}_{t-1}^{(\mathbf{D})}, \mathbf{H}_i^{(\mathbf{D})} \rangle,
\end{equation}
where $\mathbf{H}_i^{(\mathbf{D})}$ represents the encoder output at time step $i$.

To incorporate locality information, a Gaussian temporal prior centered at $p_t$ is defined as:
\begin{equation}
G(i, p_t) = \exp\left(-\frac{(i - p_t)^2}{2\sigma^2}\right),
\end{equation}
where $\sigma$ controls the width of the local attention window.

The final attention distribution is obtained by combining content relevance and the temporal prior in a normalized form:
\begin{equation}
\alpha_{t,i}^{(\mathbf{D})} =
\frac{
\exp(e_{t,i}) \cdot G(i, p_t)
}{
\sum_{j=1}^{L} \exp(e_{t,j}) \cdot G(j, p_t)
}.
\end{equation}

This formulation enables the model to focus on a temporally constrained neighborhood while preserving content-based relevance, which is particularly suitable for modeling high-frequency wind gust fluctuations.

\subsubsection{Decoder and cross-frequency modulation}\label{sec:Decoder}

The decoder generates future wind gust values in an autoregressive manner. At each decoding step $t$, the context vector is first obtained by aggregating the encoder outputs with the attention weights:
\begin{equation}
\mathbf{c}_t^{(b)} = \sum_{i=1}^{L} \alpha_{t,i}^{(b)} \mathbf{H}_i^{(b)},
\end{equation}
where $b \in \{\mathbf{A}, \mathbf{D}\}$ denotes the frequency branch.

The decoder then updates its hidden state as
\begin{equation}
\mathbf{s}_t^{(b)} =
\mathrm{LSTM}
\big(
[\phi_{\mathrm{in}}(\hat{y}_{t-1}^{(b)});\mathbf{c}_t^{(b)}],
\mathbf{s}_{t-1}^{(b)}
\big),
\end{equation}
where $\hat{y}_{t-1}^{(b)}$ denotes the previous prediction, $\phi_{\mathrm{in}}(\cdot)$ is a linear embedding layer mapping the scalar input into a $2H$-dimensional representation, and $[\cdot;\cdot]$ denotes concatenation. The decoder hidden and cell states are initialized from the corresponding bidirectional encoder outputs.

To enable information exchange between frequency components, a FiLM-based cross-frequency modulation mechanism is introduced after the decoder state update. Specifically, the low-frequency branch generates adaptive modulation parameters for the high-frequency representation:
\begin{equation}
[\boldsymbol{\gamma}_t,\boldsymbol{\beta}_t]
=
f_{\theta}(\mathbf{m}_t^{(\mathbf{A})}),
\end{equation}
where $f_{\theta}(\cdot)$ denotes a multilayer perceptron, and $\mathbf{m}_t^{(\mathbf{A})}$ represents the intermediate decoder feature of the low-frequency branch.

The high-frequency feature is then recalibrated as
\begin{equation}
\tilde{\mathbf{m}}_t^{(\mathbf{D})}
=
\boldsymbol{\gamma}_t
\odot
\mathbf{m}_t^{(\mathbf{D})}
+
\boldsymbol{\beta}_t ,
\end{equation}
where $\odot$ denotes element-wise multiplication. This modulation enables the high-frequency branch to adaptively adjust its response according to the low-frequency atmospheric evolution state.

Finally, the branch outputs are obtained through linear projections:
\begin{equation}
\hat{y}_t^{(\mathbf{A})}
=
\mathbf{W}_{\mathrm{out}}^{(\mathbf{A})}
[\mathbf{s}_t^{(\mathbf{A})};\mathbf{c}_t^{(\mathbf{A})}],
\end{equation}

\begin{equation}
\hat{y}_t^{(\mathbf{D})}
=
\mathbf{W}_{\mathrm{out}}^{(\mathbf{D})}
[\tilde{\mathbf{m}}_t^{(\mathbf{D})};\mathbf{c}_t^{(\mathbf{D})}],
\end{equation}

where $\mathbf{W}_{\mathrm{out}}^{(b)} \in \mathbb{R}^{1\times4H}$.

\subsubsection{Output reconstruction}\label{sec:Output reconstruction}

The final wind gust prediction is obtained by summing the outputs of the two branches:
\begin{equation}
\hat{y}_t = \hat{y}_t^{(\mathbf{A})} + \hat{y}_t^{(\mathbf{D})}.
\end{equation}

This additive formulation reflects the decomposition of wind gust dynamics into low-frequency trend and high-frequency fluctuation components via the stationary wavelet transform. By separately modeling each component and combining their predictions, the model captures both large-scale atmospheric evolution and transient local variations in a complementary manner.

\section{Tables}\label{sec5}

\begin{table}[h]
\centering
\caption{RMSE$\downarrow$ results of different models for WGF}
\label{tab:rmse}  
\begin{tabular}{lcccccccccc}
\toprule
Models
& 0h  & 1h & 2h & 3h & 6h & 9h & 12h & 18h & 21h  & 23h \\
\midrule
CNN-LSTM        
& 0.91 & 1.25 & 1.52 & 1.72 & 2.13 & 2.39 & 2.61 & 2.90 & 2.99 & 3.05\\
Autoformer      
& 1.78 & 1.96 & 2.10 & 2.20 & 2.44 & 2.70 & 2.89 & 3.15 & 3.23 & 3.44\\
TimesNet        
& 0.97 & 1.30 & 1.56 & 1.76 & 2.21 & 2.48 & 2.69 & 3.06 & 3.21 & 3.29\\
iTransformer    
& 0.89 & 1.24 & 1.49 & 1.68 & 2.07 & 2.32 & 2.50 & 2.77 & 2.89 & 2.97\\
PatchTST        
& 0.89 & 1.30 & 1.59 & 1.82 & 2.29 & 2.57 & 2.77 & 3.01 & 3.06 & 3.12\\
DLinear         
& 0.88 & 1.32 & 1.64 & 1.88 & 2.31 & 2.54 & 2.71 & 2.87 & 2.90 & 2.95\\
ECMWF           
& 1.93 & 1.81 & 1.83 & 1.94 & 2.07 & \textbf{1.72} & \textbf{1.78} & \textbf{1.67} & \textbf{1.85} & \textbf{1.79}\\
Ours            
& \textbf{0.84} & \textbf{1.21} & \textbf{1.46} & \textbf{1.64} & \textbf{1.98} &\textbf{2.21} &  \textbf{2.39} &  \textbf{2.63} &  \textbf{2.72} &  \textbf{2.79}\\

\bottomrule
\end{tabular}
\footnotesize{0\,h, 1\,h to 23\,h represent the forecast horizons, with 0\,h indicating the first prediction step based on the previous 24-hour input sequence. $\downarrow$ indicates that lower values are better.}
\end{table}

\begin{table}[h]
\centering
\caption{MAE$\downarrow$ results of different models for WGF}
\label{tab:mae}  
\begin{tabular}{lcccccccccc}
\toprule
Models
& 0h  & 1h & 2h & 3h & 6h & 9h & 12h & 18h & 21h  & 23h \\
\midrule
CNN-LSTM        
& 0.62 & 0.88 & 1.07 & 1.20 & 1.47 & 1.63 & 1.75 & 1.93 & 2.00 & 2.06\\
Autoformer      
& 1.22 & 1.34 & 1.44 & 1.51 & 1.66 & 1.84 & 2.02 & 2.24 & 2.30 & 2.51 \\
TimesNet        
& 0.66 & 0.90 & 1.08 & 1.22 & 1.51 & 1.69 & 1.83 & 2.04 & 2.15 & 2.22\\
iTransformer    
& 0.60 & 0.87 & 1.05 & 1.18 & 1.43 & 1.60 & 1.71 & 1.87 & 1.95 & 2.03\\
PatchTST    
& 0.60 & 0.91 & 1.13 & 1.29 & 1.60 & 1.78 & 1.88 & 2.00 & 2.02 & 2.05\\
DLinear         
& 0.57 & 0.92 & 1.17 & 1.35 & 1.67 & 1.82 & 1.92 & 2.01 & 2.01 & 2.04 \\
ECMWF           
& 1.52 & 1.36 & 1.35 & 1.43 & 1.50 & \textbf{1.27} & \textbf{1.31} & \textbf{1.24} & \textbf{1.36} & \textbf{1.29}\\

Ours            
& \textbf{0.56} & \textbf{0.85} & \textbf{1.03} & \textbf{1.16} & \textbf{1.39} &  \textbf{1.53} & \textbf{1.63} & \textbf{1.76} & \textbf{1.81} & \textbf{1.85}\\

\bottomrule
\end{tabular}
\footnotesize{0\,h, 1\,h to 23\,h represent the forecast horizons, with 0\,h indicating the first prediction step based on the previous 24-hour input sequence.}
\end{table}

\begin{table}[h]
\centering
\caption{Relative accuracy (RAcc)$\uparrow$ results of different models for WGF}
\label{tab:racc}  
\begin{tabular}{lcccccccccc}
\toprule
Models
& 0h  & 1h & 2h & 3h & 6h & 9h & 12h & 18h & 21h  & 23h \\
\midrule
CNN-LSTM        
& 0.90 & \textbf{0.87} & \textbf{0.84} & \textbf{0.82} & 0.78 & 0.75 & 0.73 & 0.70 & 0.69 & 0.68\\
Autoformer      
& 0.77 & 0.74 & 0.72 & 0.71 & 0.68 & 0.65 & 0.61 & 0.56 & 0.54 & 0.50\\
TimesNet        
& 0.88 & 0.83 & 0.79 & 0.77 & 0.73 & 0.70 & 0.67 & 0.63 & 0.61 & 0.60\\
iTransformer    
& 0.89 & 0.84 & 0.80 & 0.78 & 0.73 & 0.70 & 0.68 & 0.66 & 0.63 & 0.62\\
PatchTST        
& 0.89 & 0.83 & 0.79 & 0.76 & 0.71 & 0.68 & 0.66 & 0.64 & 0.64 & 0.63\\
DLinear         
& 0.90 & 0.83 & 0.78 & 0.73 & 0.66 & 0.62 & 0.59 & 0.58 & 0.59 & 0.58\\
ECMWF           
& 0.74 & 0.80 & 0.81 & 0.81 & \textbf{0.81} & \textbf{0.80} & \textbf{0.73} & \textbf{0.74} & \textbf{0.71} & \textbf{0.76}\\
Ours            
& \textbf{0.90} & \textbf{0.84} & \textbf{0.81} & \textbf{0.78} & \textbf{0.74} & \textbf{0.72} & \textbf{0.71} & \textbf{0.69} & \textbf{0.68} & \textbf{0.67}\\

\bottomrule
\end{tabular}
\footnotesize{0\,h, 1\,h to 23\,h represent the hours from 00:00 to 23:00, with 0\,h being the first hour predicted based on the input historical data. $\uparrow$ indicates that higher values are better.}
\end{table}

\begin{table}[h]
\centering
\caption{Sensitivity analysis of wavelet basis and decomposition level.}
\label{tab:ablation_wavelet}  
\begin{tabular}{lcccccc}
\toprule
Metric & Haar ($m{=}1$) & db2 ($m{=}1$) & db4 ($m{=}1$) & sym4 ($m{=}1$) & sym4 ($m{=}2$) & sym4 ($m{=}3$) \\
\midrule
RMSE$\downarrow$ & 2.47 ± 0.31 & 2.38 ± 0.11 & 2.41 ± 0.12 & \textbf{2.28 ± 0.03} & 2.59 ± 0.58 & 2.51 ± 0.42 \\
MAE$\downarrow$  & 1.65 ± 0.20 & 1.59 ± 0.05 & 1.60 ± 0.06 & \textbf{1.55 ± 0.04} & 1.74 ± 0.46 & 1.69 ± 0.31  \\
RAcc$\uparrow$    & 0.72 ± 0.02 & 0.72 ± 0.03 & 0.72 ± 0.03 & \textbf{0.73 ± 0.01} & 0.71 ± 0.04 & 0.71 ± 0.03 \\
\bottomrule
\end{tabular}
\vspace{2pt}
\footnotesize{
Different wavelet bases are compared under a fixed decomposition level ($m=1$), while the effect of decomposition depth is evaluated using the sym4 wavelet. }
\end{table}

\begin{table}[h]
\centering
\caption{Ablation study on model components.}
\label{tab:ablation_components}  
\setlength{\tabcolsep}{4pt}
\begin{tabular}{lcccccccc}
\toprule
Metric & w/o SWT & w/o FiLM & fixed SWT & Only A & Only D & Only global & Only local & Ours \\
\midrule
RMSE↓ & 2.92 ± 0.10 & 2.30 ± 0.06 & 2.38 ± 0.09 & 2.33 ± 0.04 & 2.32 ± 0.03 & 2.34 ± 0.05 & 2.39 ± 0.15 & \textbf{2.28 ± 0.03} \\
MAE↓  & 1.89 ± 0.05 & \textbf{1.53 ± 0.04} & 1.61 ± 0.06 & 1.56 ± 0.03 & 1.54 ± 0.01 & 1.60 ± 0.09 & 1.60 ± 0.16 & \textbf{1.55 ± 0.04} \\
RAcc↑    & 0.73 ± 0.02 & 0.72 ± 0.01 & 0.71 ± 0.04 & 0.72 ± 0.02 & 0.72 ± 0.01 & 0.69 ± 0.04 & 0.70 ± 0.05 & \textbf{0.73 ± 0.01} \\
\bottomrule
\end{tabular}
\vspace{2pt}
\footnotesize{
w/o SWT: without wavelet decomposition; w/o FiLM: without FiLM-based cross-frequency modulation; fixed SWT: stationary wavelet transform with fixed (non-trainable) filters; Only A: approximation (low-frequency) branch only; Only D: detail (high-frequency) branch only; Only global/local: using only global or local attention; Ours: full model with learnable SWT, FiLM-based cross-frequency modulation, dual-branch structure, and hybrid attention. }
\end{table}

\clearpage
\section{Figures}

\backmatter

\clearpage
\bmhead{Supplementary information}
We have Supplementary information.

\bmhead{Acknowledgements}
We extend our sincere gratitude to the researchers at ECMWF for their
invaluable contributions to the collection, archival, dissemination, and
maintenance of the ERA5 reanalysis dataset and ECMWF HRES forecast data.
\section*{Declarations}

\begin{itemize}
\item Funding

Not applicable.
\item Competing interests

The authors declare no competing interests.
\item Ethics approval and consent to participate

Not applicable.
\item Consent for publication

Not applicable.
\item Data availability  

ERA5 reanalysis data were obtained from the Copernicus Climate Data Store at 
\url{https://cds.climate.copernicus.eu/datasets/reanalysis-era5-single-levels?tab=download}. ECMWF-HRES forecast data were obtained from the European Centre for Medium-Range Weather Forecasts at  
\url{https://www.ecmwf.int/en/forecasts/datasets/set-i}. The tropical cyclone best-track data used for generating the typhoon track schematic map can be accessed at \url{https://www.ncei.noaa.gov/products/international-best-track-archive}.
\item Materials availability

Not applicable.
\item Code availability 

The code for the comparison algorithm was from public content in
GitHub: \url{https://github.com/thuml/Time-Series-Library}.
\item Author contribution

X. W performed data curation, developed the methodology, conducted the formal analyses, and wrote the original draft of the manuscript. T. L, H. Z and S. Z conceptualized the study, supervised the research, and contributed to the review and editing of the manuscript. All authors discussed the results and reviewed and approved the final manuscript.
\end{itemize}

%%===========================================================================================%%
%% If you are submitting to one of the Nature Portfolio journals, using the eJP submission   %%
%% system, please include the references within the manuscript file itself. You may do this  %%
%% by copying the reference list from your .bbl file, paste it into the main manuscript .tex %%
%% file, and delete the associated \verb+\bibliography+ commands.                            %%
%%===========================================================================================%%

\bibliography{main}% common bib file
%% if required, the content of .bbl file can be included here once bbl is generated
%%\input sn-article.bbl

\end{document}